\documentclass[10pt,twocolumn,letterpaper]{article}

\usepackage{iccv}
\usepackage{times}
\usepackage{epsfig}
\usepackage{graphicx}
\usepackage{amsmath}
\usepackage{amssymb}
\usepackage{booktabs}
\usepackage{multibib}
\usepackage{xcolor,colortbl}
\usepackage{caption}
\usepackage[pagebackref=true,breaklinks=true,letterpaper=true,colorlinks,bookmarks=false]{hyperref}

\iccvfinalcopy % *** Uncomment this line for the final submission

\def\iccvPaperID{4886} % *** Enter the ICCV Paper ID here

\DeclareUnicodeCharacter{2702}{ }
\DeclareUnicodeCharacter{FE0F}{ }

\makeatletter
\renewcommand{\paragraph}{%
  \@startsection{paragraph}{4}%
  {\z@}{0ex \@plus 1ex \@minus .2ex}{-1em}%
  {\normalfont\normalsize\bfseries}%
}
\makeatother

\usepackage{overpic}
\usepackage{enumitem} %< control spacing in itemize/enumerate/...
\usepackage{overpic} %< add raw math symbols to figures
\usepackage{color}
\usepackage{tabularx}
\definecolor{turquoise}{cmyk}{0.65,0,0.1,0.3}
\definecolor{purple}{rgb}{0.65,0,0.65}
\definecolor{dark_green}{rgb}{0, 0.5, 0}
\definecolor{orange}{rgb}{0.8, 0.6, 0.2}
\definecolor{red}{rgb}{0.8, 0.2, 0.2}
\definecolor{darkred}{rgb}{0.6, 0.1, 0.05}
\definecolor{blueish}{rgb}{0.3, 0.3, .6}
\definecolor{light_gray}{rgb}{0.7, 0.7, .7}
\definecolor{pink}{rgb}{1, 0, 1}
\definecolor{greyblue}{rgb}{0.25, 0.25, 1}
\definecolor{awesome}{rgb}{1.0, 0.13, 0.32}

\definecolor{figred}{rgb}{0.9, 0.1, 0.1}
\definecolor{figgreen}{rgb}{0.1, 0.7, 0.1}
\definecolor{figblue}{rgb}{0.1, 0.1, 0.9}
\definecolor{figmagenta}{rgb}{0.8, 0.1, 0.8}

\usepackage{blindtext}

\renewcommand{\paragraph}[1]{\vspace{1em}\noindent\textbf{#1}}

\usepackage{gensymb}

\usepackage{graphicx}               % Necessary to use \scalebox

\usepackage{tabu}
\usepackage{xcolor}

\begin{document}

%%%%%%%%% TITLE
% \title{Turn-the-Cam: Learning to Control the Camera in Large Diffusion Models}
% \title{Zero-Shot Control of Camera Viewpoints from a Single Image}
\title{Zero-Shot Control of Camera Viewpoint within a Single Image}
\title{Zero-Shot Control of a Single Image's Viewpoint}
\title{Zero-Shot Control of the Viewpoint of a Single Image}
\title{Zero-Shot Control of the Image Viewpoint}
\title{Zero-shot View Synthesis from a Single Image}
\title{Zero-1-to-3: Zero-shot One Image to 3D}
\title{Turn-the-Cam: Turning Camera Viewpoint with One Image}
\title{Zero-1-to-3: Zero-shot One Image to 3D Object}
% \title{Zero-1-to-3 \\ Zero-shot One Image to 3D for Novel View Synthesis and 3D Reconstruction}
%\title{1-2-3 from Zero: One 2D Image to 3D with Zero-shots}
% \title{CONCAT: CONtrolling CAmera exTrinsics in Large Diffusion Models}

\author{Ruoshi Liu$^1$ \ \ Rundi Wu$^1$ \ \ Basile Van Hoorick$^1$ \ \ Pavel Tokmakov$^2$ \ \ Sergey Zakharov$^2$ \ \ Carl Vondrick$^1$ 
\vspace{0.1cm}
\\$^1$
\hspace{.1cm}Columbia University \ \ $^2$\hspace{.1cm}Toyota Research Institute
\vspace{0.08cm}
\\
%{\small{\tt\{rliu,rundi,basile,vondrick\}@cs.columbia.edu}\hspace{.5cm}{\tt\{pavel.tokmakov,sergey.zakharov\}@tri.global}} \\
  \href{https://zero123.cs.columbia.edu/}{\textbf{\url{zero123.cs.columbia.edu}}}
% \vspace{-0.32cm}
}

\twocolumn[{%
\renewcommand\twocolumn[1][]{#1}%
\maketitle
\begin{center}
\vspace{-0.18cm}
        \includegraphics[width=0.98\linewidth]{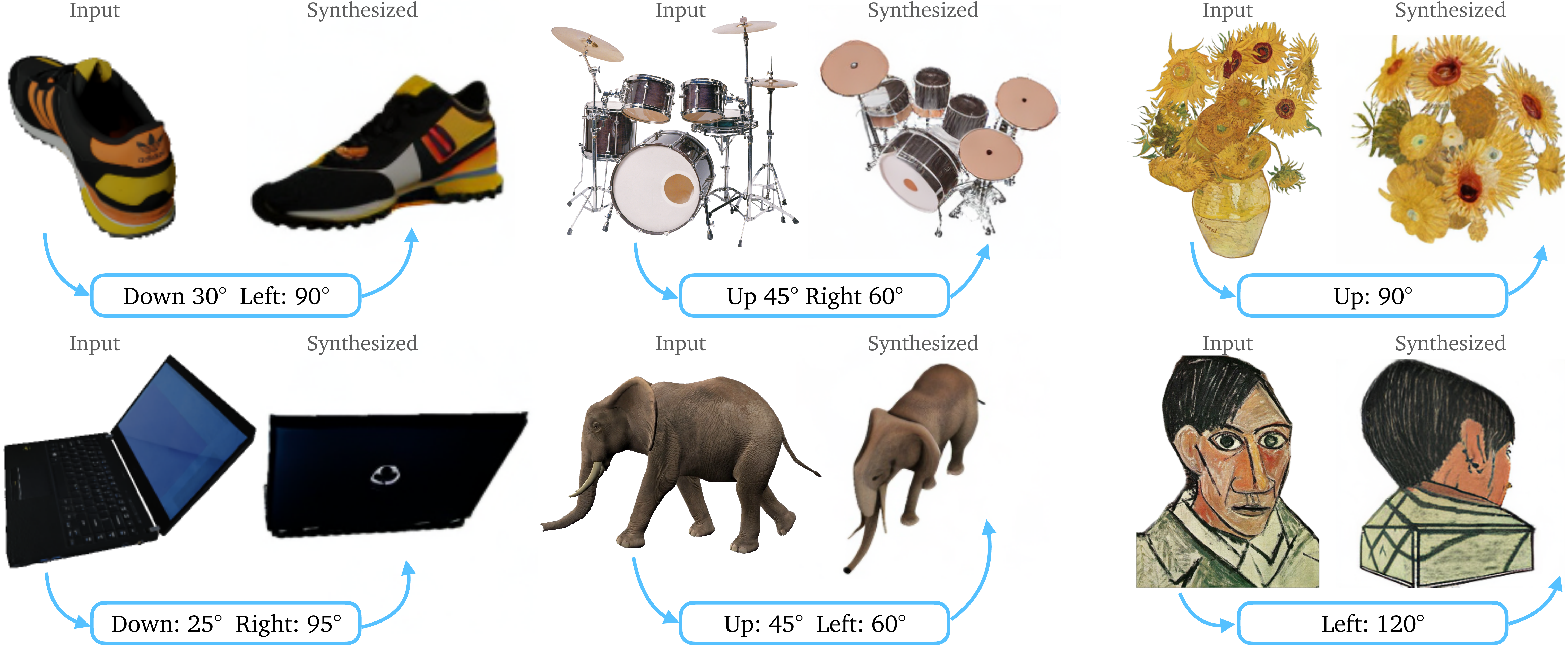}
        \captionof{figure}{Given a single RGB image of an object, we present \textbf{\textit{Zero-1-to-3}}, a method  to synthesize an image from a specified camera viewpoint. Our approach synthesizes views that contain rich details consistent with the input view for large relative transformations. It also achieves strong zero-shot performance on objects with complex geometry and artistic styles.
        }
%        \vspace{+0.5cm}
        \label{fig:teaser}
\end{center}
}]


%%%%%%%%% ABSTRACT

\begin{abstract}
%\vspace{-0.5cm}
%Given a single image of an object, we present a method, \textbf{Zero-1-to-3}, to generate another image of the same object from any viewpoint. Large-scale pretrained diffusion models contain rich semantic and geometric priors about objects, but they lack a way to extract and control. To this end, we create a synthetic dataset with \textit{Objaverse} and fine-tune \textit{Stable Diffusion} to generate novel views, conditioned on an input view and the relative viewpoint transformation. Our viewpoint-conditioned diffusion approach can further be used for the task of 3D reconstruction from a single image. Even though it is trained on a synthetic dataset, our model retains a strong generalization ability to out-of-distribution datasets as well as in-the-wild images, including impressionist paintings. Qualitative and quantitative experiments show that our method significantly outperforms state-of-the-art single-view 3D reconstruction and novel view synthesis models by capitalizing on internet-scale pertaining.

%\vspace{-2em}
We introduce Zero-1-to-3, a framework for changing the camera viewpoint of an object given just a single RGB image. To perform novel view synthesis in this under-constrained setting, we capitalize on the geometric priors that large-scale diffusion models learn about natural images. Our conditional diffusion model uses a synthetic dataset to learn controls of the relative camera viewpoint, which allow new images to be generated of the same object under a specified camera transformation. Even though it is trained on a synthetic dataset, our model retains a strong zero-shot generalization ability to out-of-distribution datasets as well as in-the-wild images, including impressionist paintings. Our viewpoint-conditioned diffusion approach can further be used for the task of 3D reconstruction from a single image. Qualitative and quantitative experiments show that our method significantly outperforms state-of-the-art single-view 3D reconstruction and novel view synthesis models by leveraging Internet-scale pre-training.
\end{abstract}

\vspace{-3.8em}
%%%%%%%%% BODY TEXT

\section{Introduction}
\vspace{+0.25cm}
% Large-scale diffusion models contain severer viewpoint bias. When prompted to generate images of chairs, it's extremely unlikely the generated images contain a chair viewed from its back. Recent work in text-to-3D and image-to-3D has suffered from this bias, known as the "Janus Problem". To generate the 3D shape of "a chair" from text, or reconstruct a chair from an image, these methods optimize a NeRF (or its variants) with a languaged-conditioned diffusion model.

% We address the challenging dual problems of zero-shot novel view synthesis and 3D reconstruction from a single uncalibrated RGB image. Due eometr severely under-constrained nature of the problems, we need to rely on both semantic, and geometric priors. Though it's impossible to train a large-scale 3D generative models with an internet-scale 3D dataset today, we believe large diffusion models like \textit{Stable Diffusion}~\cite{rombach2022high}, trained to generate 

%\looseness=-1
From just a single camera view, humans are often able to imagine an object's 3D shape and appearance. This ability is important for everyday tasks, such as object manipulation~\cite{gupta2009observing} and navigation in complex environments~\cite{cheng2005reflections}, but is also key for visual creativity, such as painting~\cite{morriss2010evolution}. While this ability can be partially explained by reliance on geometric priors like symmetry, we seem to be able to generalize to much more challenging objects that break physical and geometric constraints with ease. In fact, we can predict the 3D shape of objects that do not (or even \emph{cannot}) exist in the physical world (see third column in Figure~\ref{fig:teaser}). To achieve this degree of generalization, humans rely on prior knowledge accumulated through a lifetime of visual exploration.

In contrast, most existing approaches for 3D image reconstruction operate in a closed-world setting due to their reliance on expensive 3D annotations (e.g.\ CAD models) or category-specific priors~\cite{pavlakos2019expressive,huang2022planes,park2019deepsdf,zuffi2018lions,Zuffi:CVPR:2017, zuffi2019three,kanazawa2019learning,kanazawa2018learning}. Very recently, several methods have made major strides in the direction of open-world 3D reconstruction by pre-training on large-scale, diverse datasets such as CO3D \cite{reizenstein2021common,mescheder2019occupancy,park2019deepsdf,gao2022get3d}. However, these approaches often still require geometry-related information for training, such as stereo views or camera poses. As a result, the scale and diversity of the data they use remain insignificant compared to the recent Internet-scale text-image collections~\cite{schuhmann2022laion} that enable the success of large diffusion models~\cite{saharia2022photorealistic,rombach2022high,nichol2021glide}. It has been shown that Internet-scale pre-training endows these models with rich semantic priors, but the extent to which they capture geometric information remains largely unexplored.

In this paper, we demonstrate that we are able to learn control mechanisms that manipulate the camera viewpoint in large-scale diffusion models, such as Stable Diffusion \cite{rombach2022high}, in order to perform zero-shot novel view synthesis and 3D shape reconstruction. Given a single RGB image, both of these tasks are severely under-constrained. However, due to the scale of training data available to modern generative models (over 5 billion images), diffusion models are state-of-the-art representations for the natural image distribution, with support that covers a vast number of objects from many viewpoints. Although they are trained on 2D monocular images without any camera correspondences, we can fine-tune the model to learn controls for relative camera rotation and translation during the generation process. These controls allow us to encode arbitrary images that are decoded to a different camera viewpoint of our choosing. Figure~\ref{fig:teaser} shows several examples of our results.

The primary contribution of this paper is to demonstrate that large diffusion models have learned rich 3D priors about the visual world, even though they are only trained on 2D images. We also demonstrate state-of-the-art results for novel view synthesis and state-of-the-art results for zero-shot 3D reconstruction of objects, both from a single RGB image. We begin by briefly reviewing related work in Section~\ref{sec:rel}. In Section~\ref{sec:meth}, we describe our approach to learn controls for camera extrinsics by fine-tuning large diffusion models. Finally, in Section~\ref{sec:exp}, we present several quantitative and qualitative experiments to evaluate zero-shot view synthesis and 3D reconstruction of geometry and appearance from a single image. We will release all code and models as well as an online demo.

\section{Related Work}
\label{sec:rel}
% \subsection{Neural Radiance Fields} \label{related:nerf}

\begin{figure}
    \centering
    \includegraphics[width=\linewidth]{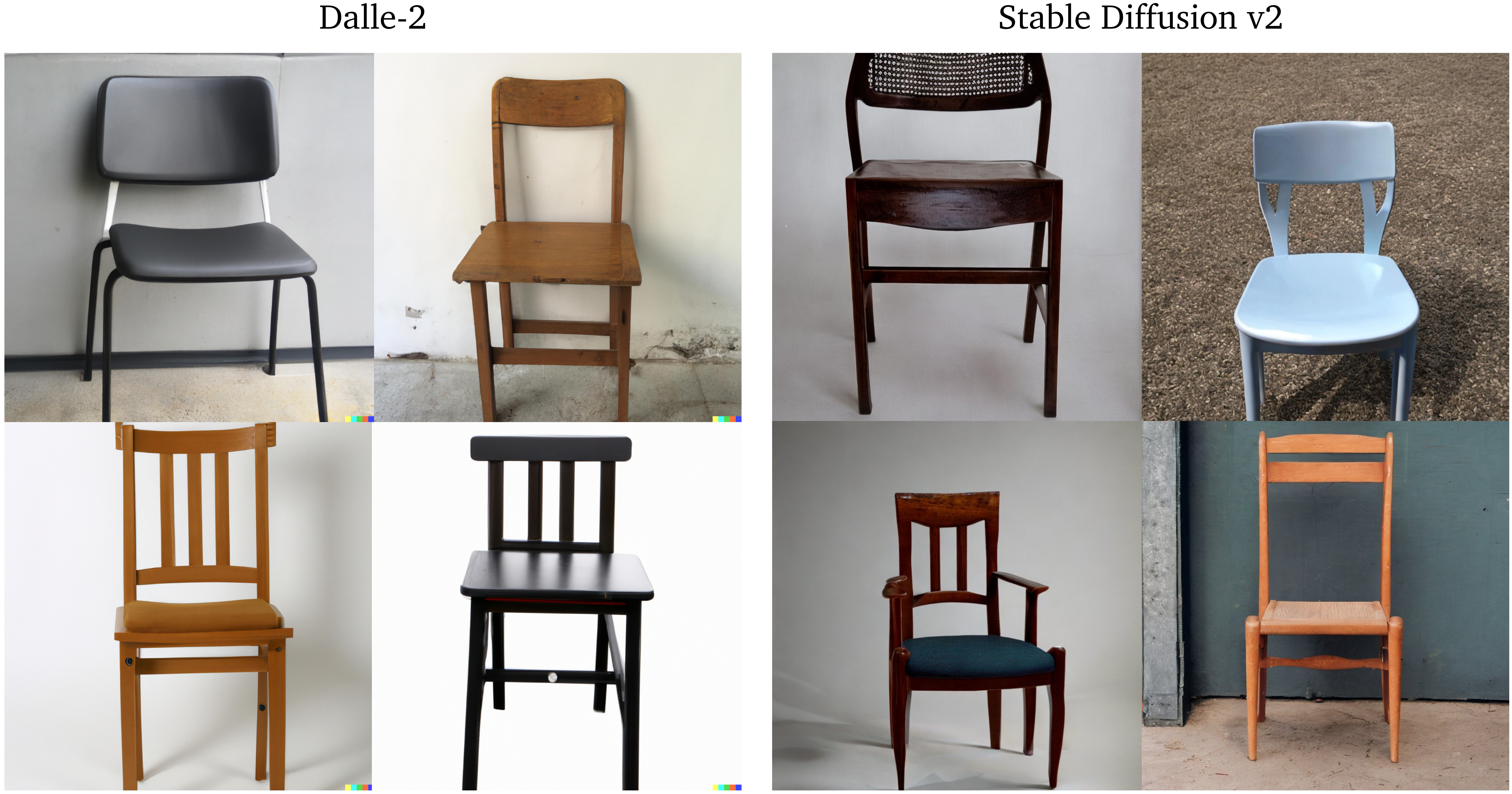}
    %\label{fig:viewpoint}
    \caption{
    \textbf{Viewpoint bias in text-to-image models.} We show samples from both Dall-E-2 and Stable Diffusion v2 from the prompt \textit{``a chair"}. Most samples show a chair in a forward-facing canonical pose.
    \vspace{-1em}
%    \vspace{-0.3cm}
    }
    \label{fig:viewpointbias}
\end{figure}

\textbf{3D generative models.}
\label{related:generative}
Recent advancements in generative image architectures combined with large scale image-text datasets~\cite{schuhmann2022laion} have made it possible to synthesize high-fidelity of diverse scenes and objects~\cite{nichol2021glide,ramesh2021zero,saharia2022photorealistic}. In particular, diffusion models have shown to be very effective at learning scalable image generators using a denoising objective~\cite{chen2020wavegrad,sohl2015deep}. However, scaling them to the 3D domain would require large amounts of expensive annotated 3D data. Instead, recent approaches rely on transferring pre-trained large-scale 2D diffusion models to 3D without using any ground truth 3D data. Neural Radiance Fields or NeRFs~\cite{mildenhall2020nerf} have emerged as a powerful representation, thanks to their ability to encode scenes with high fidelity. Typically, NeRF is used for single-scene reconstruction, where many posed images covering the entire scene are provided. The task is then to predict novel views from unobserved angles. DreamFields~\cite{jain2022zero} has shown that NeRF is a more versatile tool that can also be used as the main component in a 3D generative system. Various follow-up works~\cite{poole2022dreamfusion,lin2022magic3d,wang2022score} substitute CLIP for a distillation loss from a 2D diffusion model that is repurposed to generate high-fidelity 3D objects and scenes from text inputs.

Our work explores an unconventional approach to novel-view synthesis, modeling it as a viewpoint-conditioned image-to-image translation task with diffusion models. The learned model can also be combined with 3D distillation to reconstruct 3D shape from a single image. Prior work~\cite{watson2022novel} adopted a similar pipeline but did not demonstrate zero-shot generalization capability. Concurrent approaches~\cite{deng2022nerdi, melas2023realfusion, xu2022neurallift} proposed similar techniques to perform image-to-3D generation using language-guided priors and textual inversion~\cite{gal2022image}. In comparison, our method learns control of viewpoints through a synthetic dataset and demonstrates zero-shot generalization to in-the-wild images.

\textbf{Single-view object reconstruction.}
\label{related:single-view}
Reconstructing 3D objects from a single view is a highly challenging problem that requires strong priors. One line of work builds priors from relying on collections of 3D primitives represented as meshes~\cite{worchel2022multi,xu2019disn}, voxels~\cite{girdhar2016learning,wu2017marrnet}, or point clouds~\cite{fan2017point,mescheder2019occupancy}, and use image encoders for conditioning. These models are constrained by the variety of the used 3D data collection and show poor generalization capabilities due to the global nature of this type of conditioning. Moreover, they require an additional pose estimation step to ensure alignment between the estimated shape and the input. On the other hand, locally conditioned
models~\cite{saito2019pifu,yu2021pixelnerf,wang2021ibrnet,tucker2020single,revealing}
% models~\cite{saito2019pifu,yu2021pixelnerf,wang2021ibrnet,tucker2020single}
aim to use local image features directly for scene reconstruction and show greater cross-domain generalization capabilities, though are generally limited to close-by view reconstructions. Recently, MCC~\cite{wu2023multiview} learns a general-purpose representation for 3D reconstruction from RGB-D views and is trained on large-scale dataset of object-centric videos.

In our work, we demonstrate that rich geometric information can be extracted directly from a pre-trained Stable Diffusion model, alleviating the need for additional depth information.

% A separate body of research focuses on a problem of 3D model generation in different representations.
% Point cloud-based generative models:
% Achlioptas et al.~\cite{achlioptas2018learning} fit generative priors based on GANs~\cite{creswell2018generative} or GMMs to the latent representations of pretrained point cloud autoencoders. Mo et al. propose to use VAE~\cite{kingma2013auto} that encodes shape structure and geometry represented as a hierarchy graphs. Pointflow~\cite{yang2019pointflow} is a two-stage flow model that uses first model for latent generation, ant the second one for sampling points conditioned on the latent. Follow-up works propose to use diffusion models~\cite{luo2021diffusion,cai2020learning,zeng2022lion} at first or both stages. More recently, Point-E~\cite{nichol2022point} introduced a simple one-stage transformer-based model that also incorporates colors.

% \subsection{Large-scale 3D datasets} \label{related:dataset}

\section{Method}
\label{sec:meth}

\begin{figure}
    \centering
    \includegraphics[width=\linewidth]{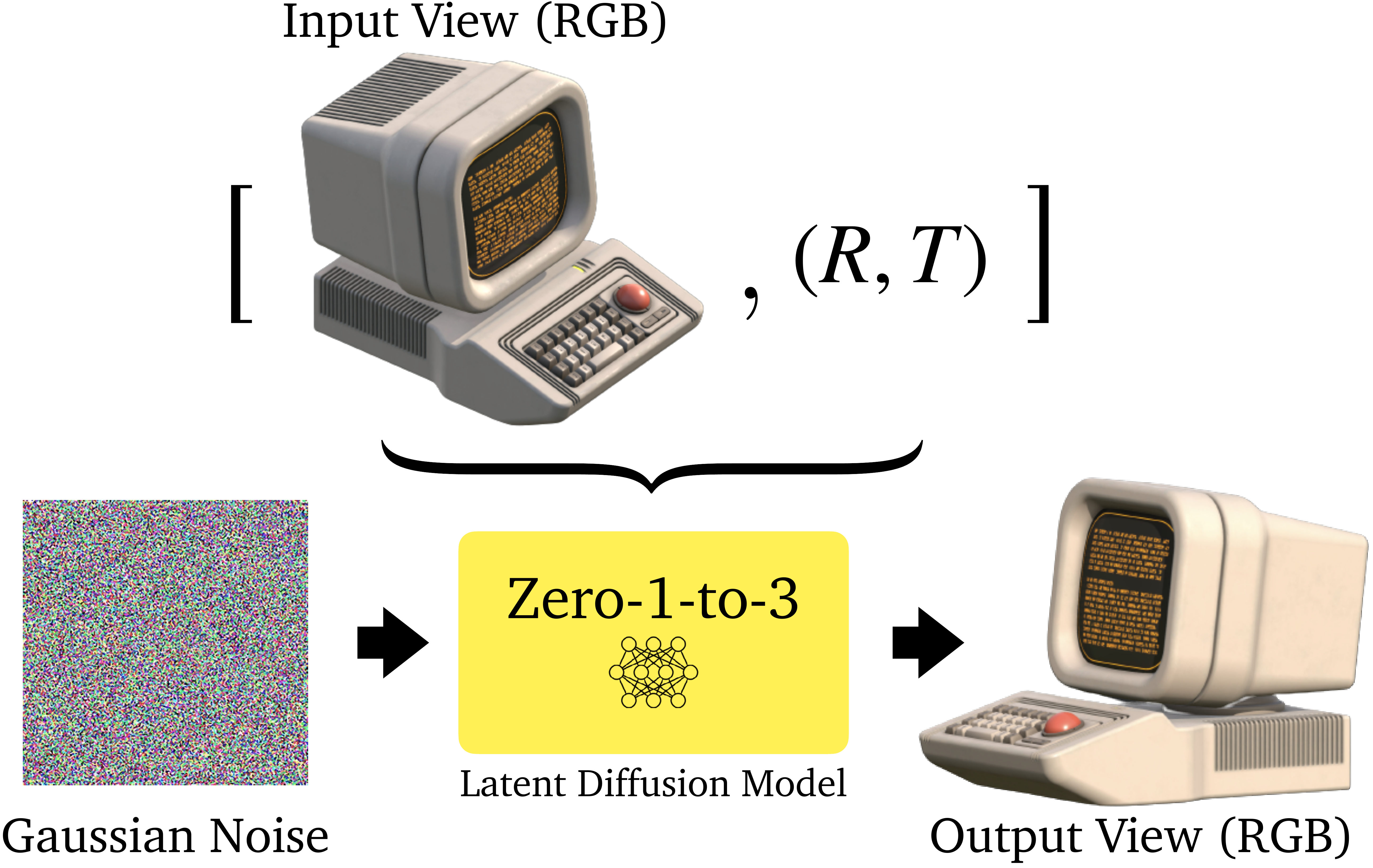}
    \caption{\textbf{Zero-1-to-3} is a viewpoint-conditioned image translation model using a conditional latent diffusion architecture. Both the input view and a relative viewpoint transformation are used as conditional information.
    % \vspace{-0.3cm}
    }
    \label{fig:method_nvs}
\end{figure}

Given a single RGB image $x \in \mathbb{R}^{H \times W \times 3}$ of an object, our goal is to synthesize an image of the object from a different camera viewpoint. Let $R \in \mathbb{R}^{3\times 3}$ and $T \in \mathbb{R}^3$ be the relative camera rotation and translation of the desired viewpoint, respectively. We aim to learn a model $f$ that synthesizes a new image under this camera transformation:
\begin{align}
    \hat{x}_{R,T} = f(x, R, T)
\end{align}
where we denote $\hat{x}_{R,T}$ as the synthesized image. We want our estimated $\hat{x}_{R,T}$  to be perceptually similar to the true but unobserved novel view $x_{R,T}$. 

Novel view synthesis from monocular RGB image is severely under-constrained. Our approach will capitalize on large diffusion models, such as Stable Diffusion, in order to perform this task, since they show extraordinary zero-shot abilities when generating diverse images from text descriptions. Due to the scale of their training data~\cite{schuhmann2022laion}, pre-trained diffusion models are state-of-the-art representations for the natural image distribution today.  

However, there are two challenges that we must overcome to create $f$. Firstly, although large-scale generative models are trained on a large variety of objects in different viewpoints, the representations do not explicitly encode the correspondences between viewpoints. Secondly, generative models inherit viewpoint biases reflected on the Internet. As shown in Figure
\ref{fig:viewpointbias}, Stable Diffusion tends to generate images of forward-facing chairs in canonical poses. These two problems greatly hinder the ability to extract 3D knowledge from large-scale diffusion models.

\begin{figure}
    % TODO: adjust figure text, it's not a neural field
    \centering
    \includegraphics[width=\linewidth]{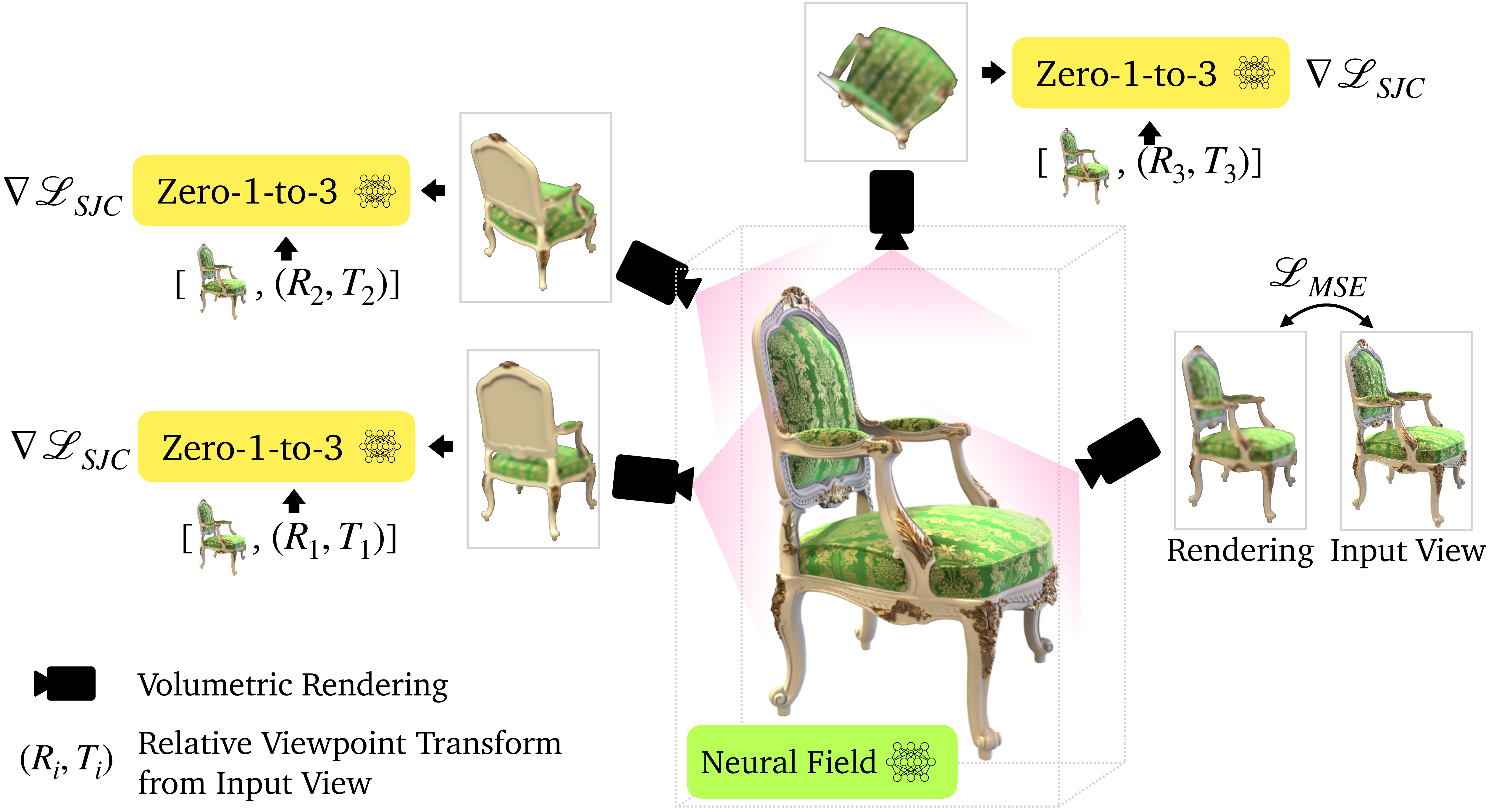}
    \caption{\textbf{3D reconstruction with Zero-1-to-3.} Zero-1-to-3 can be used to optimize a neural field for the task of 3D reconstruction from a single image. During training, we randomly sample viewpoints and use Zero-1-to-3 to supervise the 3D reconstruction.
    % \vspace{-0.3cm}
    }
    \label{fig:method_3d}
\end{figure}

\subsection{Learning to Control Camera Viewpoint}

\begin{figure*}
    % TODO: shoe is too dark, increase brightness
    \centering
    \includegraphics[width=\linewidth]{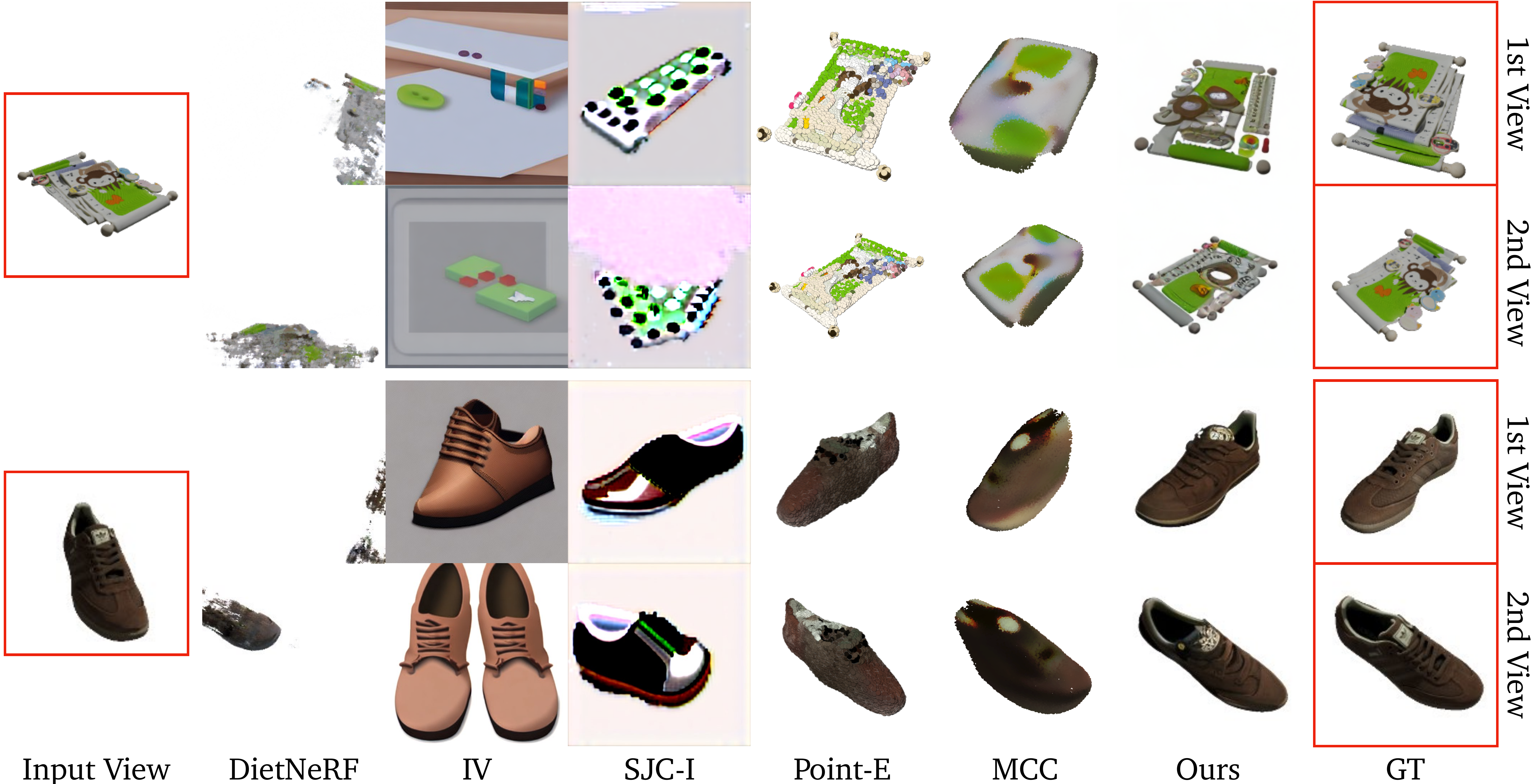}
    \caption{\textbf{Novel view synthesis on Google Scanned Objects~\cite{downs2022google}.} The input view shown on the left is used to synthesize two randomly sampled novel views. Corresponding ground truth views are shown on the right. Compared to the baselines, our synthesized novel view contain rich textual and geometric details that are highly consistent with the ground truth, while baseline methods display a significant loss of high-frequency details.
    % \vspace{-0.3cm}
    }
    \label{fig:qualitative_gso}
\end{figure*}

Since diffusion models have been trained on internet-scale data, their support of the natural image distribution likely covers most viewpoints for most objects, but these viewpoints cannot be controlled in the pre-trained models. Once we are able to \emph{teach} the model a mechanism to control the camera extrinsics with which a photo is captured, then we unlock the ability to perform novel view synthesis.

To this end, given a dataset of paired images and their relative camera extrinsics $\{ \left(x,x_{(R,T)},R,T\right) \}$, our approach, shown in Figure~\ref{fig:method_nvs}, fine-tunes a pre-trained diffusion model in order to learn controls over the camera parameters without destroying the rest of the representation. Following \cite{rombach2022high}, we use a latent diffusion architecture with an encoder $\mathcal{E}$, a denoiser U-Net $\epsilon_{\theta}$, and a decoder $\mathcal{D}$. At the diffusion time step $t \sim [1, 1000]$, let $c(x, R, T)$ be the embedding of the input view and relative camera extrinsics. We then solve for the following objective to fine-tune the model:
\begin{align}
   \min_{\theta}\;  \mathbb{E}_{z \sim \mathcal{E}(x), t, \epsilon \sim \mathcal{N}(0, 1)}||\epsilon - \epsilon_{\theta}(z_t, t, c(x, R, T))||_2^2.
\end{align}
After the model $\epsilon_\theta$ is trained, the inference model $f$ can generate an image by performing iterative denoising from a Gaussian noise image \cite{rombach2022high} conditioned on $c(x, R, T)$. 

The main result of this paper is that fine-tuning pre-trained diffusion models in this way enables them to learn a generic mechanism for controlling the camera viewpoints, which extrapolates outside of the objects seen in the fine-tuning dataset. In other words, this fine-tuning allows controls to be ``bolted on'' and the diffusion model can retain the ability to generate photorealistic images, except now with control of viewpoints. This compositionality establishes zero-shot capabilities in the model, where the final model can synthesize new views for object classes that lack 3D assets and never appear in the fine-tuning set.

\begin{figure*}
    \centering
    \includegraphics[width=\linewidth]{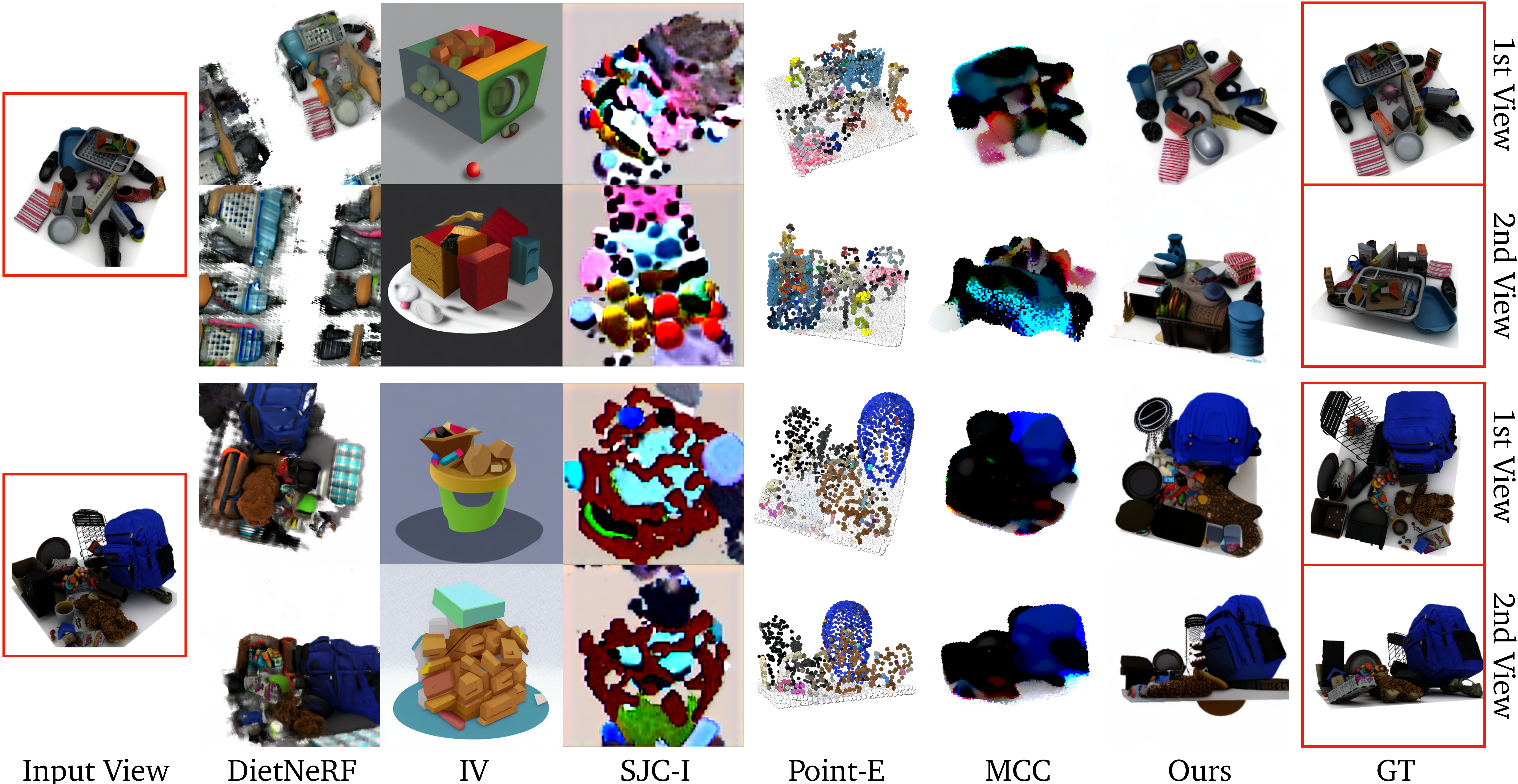}
    \caption{\textbf{Novel view synthesis on RTMV~\cite{tremblay2022rtmv}.} The input view shown on the left is used to synthesize 2 randomly sampled novel views. Corresponding ground truth views are shown on the right. Our synthesized view maintains a high fidelity even under big camera viewpoint changes, while most other methods deteriorate in quality drastically.
    % \vspace{-0.3cm}
    }
    \label{fig:qualitative_rtmv}
\end{figure*}

\subsection{View-Conditioned Diffusion}

3D reconstruction from a single image requires both low-level perception (depth, shading, texture, etc.) and high-level understanding (type, function, structure, etc.). Therefore, we adopt a hybrid conditioning mechanism. On one stream, a CLIP~\cite{radford2021learning} embedding of the input image is concatenated with $(R, T)$ to form a ``posed CLIP'' embedding $c(x, R, T)$. We apply cross-attention to condition the denoising U-Net, which provides high-level semantic information of the input image. On the other stream, the input image is channel-concatenated with the image being denoised, assisting the model in keeping the identity and details of the object being synthesized. To be able to apply classifier-free guidance~\cite{ho2022classifier}, we follow a similar mechanism proposed in~\cite{brooks2022instructpix2pix}, setting the input image and the posed CLIP embedding to a null vector randomly, and scaling the conditional information during inference.

\subsection{3D Reconstruction}

In many applications, synthesizing novel views of an object is not enough. A full 3D reconstruction capturing both the appearance and geometry of an object is desired. We adopt a recently open-sourced framework, Score Jacobian Chaining (SJC)~\cite{wang2022score}, to optimize a 3D representation with priors from text-to-image diffusion models. However, due to the probabilistic nature of diffusion models, gradient updates are highly stochastic. A crucial technique used in SJC, inspired by DreamFusion~\cite{poole2022dreamfusion}, is to set the classifier-free guidance value to be significantly higher than usual. This methodology decreases the diversity of each sample but improves the fidelity of the reconstruction.

As shown in Figure~\ref{fig:method_3d}, similarly to SJC, we randomly sample viewpoints and perform volumetric rendering. We then perturb the resulting images with Gaussian noise $\epsilon \sim \mathcal{N}(0, 1)$, and denoise them by applying the U-Net $\epsilon_{\theta}$ conditioned on the input image $x$, posed CLIP embedding $c(x, R, T)$, and timestep $t$, in order to approximate the score toward the non-noisy input $x_\pi$:
\begin{equation}
    \nabla \mathcal{L}_{SJC} = \nabla_{I_\pi} \log p_{\sqrt{2}\epsilon}(x_{\pi})
\end{equation}
where $\nabla \mathcal{L}_{SJC}$ is the PAAS score introduced by \cite{wang2022score}.

In addition, we optimize the input view with an MSE loss. To further regularize the NeRF representation, we also apply a depth smoothness loss to every sampled viewpoint, and a near-view consistency loss to regularize the change in appearance between nearby views.

\subsection{Dataset}

We use the recently released \textit{Objaverse}~\cite{deitke2022objaverse} dataset for fine-tuning, which is a large-scale open-source dataset containing 800K+ 3D models created by 100K+ artists. While it has no explicit class labels like ShapeNet~\cite{chang2015shapenet}, Objaverse embodies a large diversity of high-quality 3D models with rich geometry, many of them with fine-grained details and material properties. For each object in the dataset, we randomly sample 12 camera extrinsics matrices $\mathcal{M_e}$ pointing at the center of the object and render 12 views with a ray-tracing engine. At training time, two views can be sampled for each object to form an image pair $(x, x_{R, T})$. The corresponding relative viewpoint transformation $(R, T)$ that defines the mapping between both perspectives can easily be derived from the two extrinsic matrices.

% \BV{todo mention something about off the shelf segmentation for white background somewhere}

% \subsection{Implementation Details}

% \BV{todo mention something about architecture hyperparameters, or at least changes relative to regular LDM/SD, optimizer, learning rate, etc}

\section{Experiments}
\label{sec:exp}

We assess our model's performance on zero-shot novel view synthesis and 3D reconstruction. As confirmed by the authors of Objaverse, the datasets and images we used in this paper are outside of the Objaverse dataset, and can thus be considered zero-shot results. We quantitatively compare our model to the state-of-the-art on synthetic objects and scenes with different levels of complexity. We also report qualitative results using diverse in-the-wild images, ranging from pictures we took of daily objects to paintings.

%First, we describe the tasks and methodology in Section~\ref{sec:tasks}. In Section~\ref{sec:baselines}, we outline the baseline methods we compare our approach against. Then, we evaluate our model on novel view synthesis in Section~\ref{sec:nvs}, and on 3D reconstruction in Section~\ref{sec:3d}, both using metrics and data described in Section~\ref{sec:metrics}.

\begin{figure*}
    \centering
    \includegraphics[width=\linewidth]{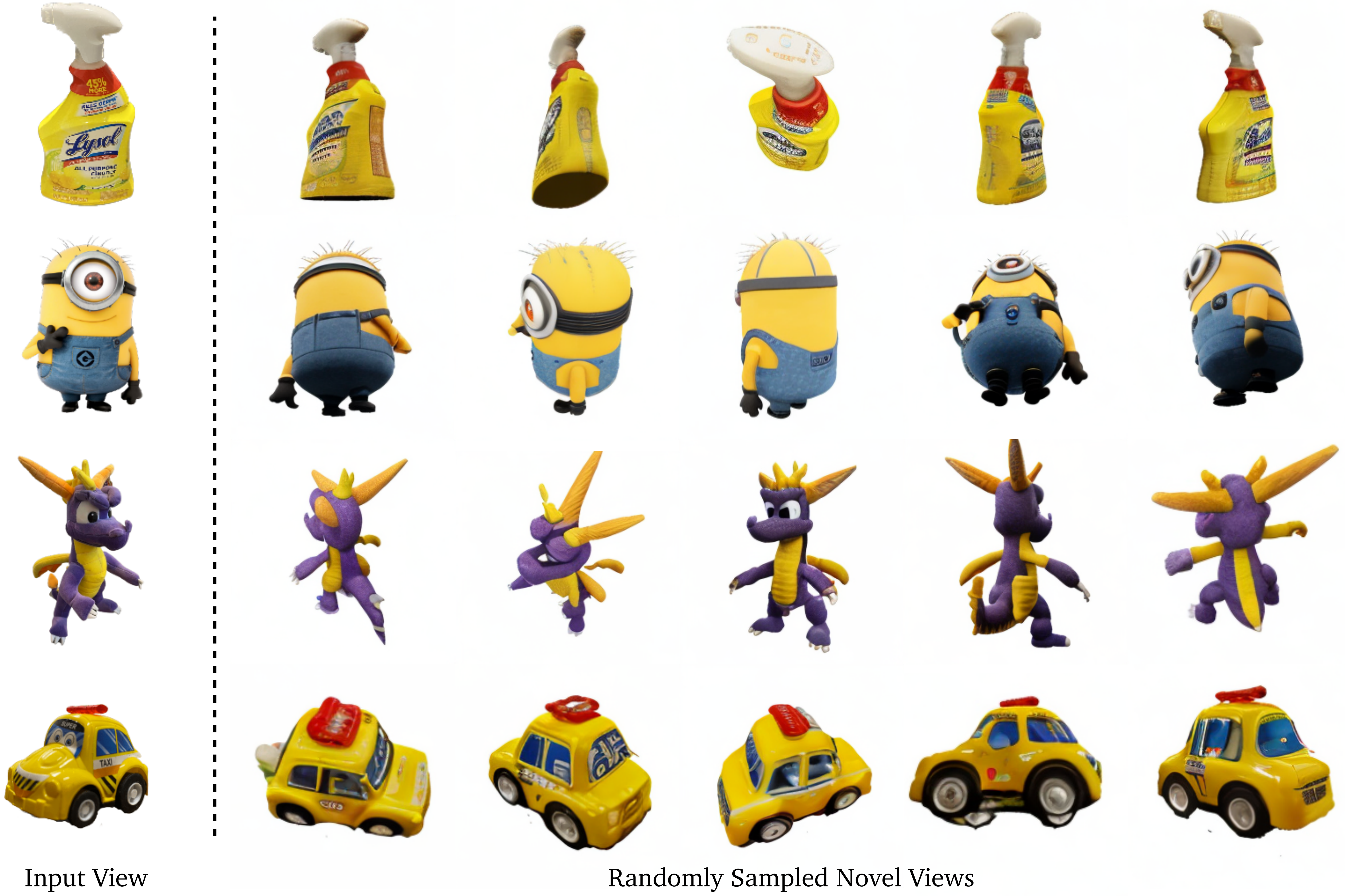}
    \caption{\textbf{Novel view synthesis on in-the-wild images.} The 1st, 3rd, and 4th rows show results on images taken by an iPhone, and the 2nd row shows results on an image downloaded from the Internet. Our method works are robust to objects with different surface materials and geometry.  We randomly sampled 5 different viewpoints and directly showcase the results without cherry-picking. We include more uncurated results in the supplementary materials.
    % \vspace{-0.3cm}
    }
    \label{fig:qualitative_wild}
\end{figure*}

\subsection{Tasks}
\label{sec:tasks}

We describe two closely related tasks that take single-view RGB images as input, and we apply them zero-shot.

\looseness=-1
\textbf{Novel view synthesis.}
Novel view synthesis is a long-standing 3D problem in computer vision that requires a model to learn the depth, texture, and shape of an object implicitly. The extremely limited input information of only a single view requires a novel view synthesis method to leverage prior knowledge. Recent popular methods have relied on optimizing implicit neural fields with CLIP consistency objectives from randomly sampled views~\cite{jain2021putting}. Our approach for view-conditional image generation is orthogonal, however, because we invert the order of 3D reconstruction and novel view synthesis, while still retaining the identity of the object depicted in the input image. This way, the aleatoric uncertainty due to self-occlusion can be modeled by a probabilistic generative model when rotating around objects, and the semantic and geometric priors learned by large diffusion models can be leveraged effectively.

%\vspace{-0.2cm}
\looseness=-1
\textbf{3D Reconstruction.}
We can also adapt a stochastic 3D reconstruction framework such as SJC~\cite{wang2022score} or DreamFusion~\cite{poole2022dreamfusion} to create a most likely 3D representation. We parameterize this as a voxel radiance field~\cite{chen2022tensorf,sun2022direct,fridovich2022plenoxels}, and subsequently extract a mesh by performing marching cubes on the density field. The application of our view-conditioned diffusion model for 3D reconstruction provides a viable path to channel the rich 2D appearance priors learned by our diffusion model toward 3D geometry.

\subsection{Baselines}
\label{sec:baselines}

To be consistent with the scope of our method, we compare only to methods that operate in a zero-shot setting and use single-view RGB images as input.

For novel view synthesis, we compare against several state-of-the-art, single-image algorithms. In particular, we benchmark DietNeRF~\cite{jain2021putting}, which regularizes NeRF with a CLIP image-to-image consistency loss across viewpoints. In addition, we compare with Image Variations (IV)~\cite{sdvariation}, which is a Stable Diffusion model fine-tuned to be conditioned on images instead of text prompts and could be seen as a semantic nearest-neighbor search engine with Stable Diffusion. Finally, we adapted SJC~\cite{wang2022score}, a diffusion-based text-to-3D model where the original text-conditioned diffusion model is replaced with an image-conditioned diffusion model, which we termed SJC-I.

For 3D reconstruction, we use two state-of-the-art, single-view algorithms as baselines:
(1) Multiview Compressive Coding (MCC)~\cite{wu2023multiview}, which is a neural field-based approach that completes RGB-D observations into a 3D representation, as well as (2) Point-E~\cite{nichol2022point}, which is a diffusion model over colorized point clouds. MCC is trained on CO3Dv2 \cite{reizenstein2021common}, while Point-E is notably trained on a significantly bigger OpenAI's internal 3D dataset. We also compare against SJC-I. % with much more computing resources.

\begin{figure*}
    \centering
    \includegraphics[width=\linewidth]{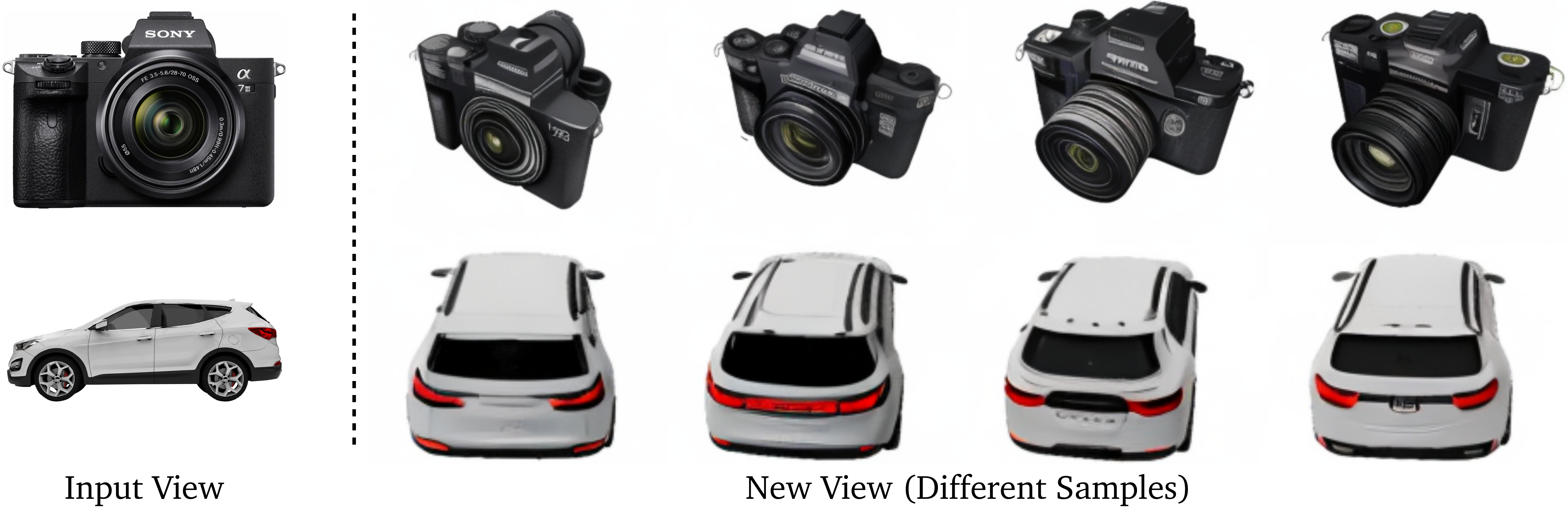}
    \caption{\textbf{Diversity of novel view synthesis.} With an input view, we fix another viewpoint and randomly generate multiple conditional samples. The different results reflect a range of diversity in terms of both geometry and appearance information that is missing in the input view.
    % \vspace{-0.3cm}
    }
    \label{fig:diversity}
\end{figure*}

Since MCC requires depth input, we use MiDaS~\cite{ranftl2020towards, ranftl2021vision} off-the-shelf  for depth estimation. % After obtaining disparity maps,
% and use \bv{todo which? should we be honest about using some random website?} an online off-the-shelf foreground segmentation model.
We convert the obtained relative disparity map to an absolute pseudo-metric depth map by assuming standard scale and shift values that look reasonable across the entire test set.

\subsection{Benchmarks and Metrics}
\label{sec:metrics}

We evaluate both tasks on Google Scanned Objects (GSO)~\cite{downs2022google}, which is a dataset of high-quality scanned household items, as well as RTMV~\cite{tremblay2022rtmv}, which consists of complex scenes, each composed of 20 random objects.
% We randomly sample 50 objects from the GSO dataset. For each such object, we reserve one viewpoint as input, and render 9 other random views to use as ground truths for evaluating novel view synthesis.
% We randomly sample 10 scenes from the RTMV dataset. For each such scene, one viewpoint is reserved as input and another 30 random views serve as ground truths for evaluating novel view synthesis.
In all experiments, the respective ground truth 3D models are used for evaluating 3D reconstruction.
% Due to limited time and resources, we use a subset of 20 objects out of the 50 GSO objects for evaluation on 3D reconstruction.

\begin{table}[t!]
    \footnotesize
    \centering
    \begin{tabular}{cccc|c}
    \toprule
    \multicolumn{1}{l}{} & DietNeRF~\cite{jain2021putting} & \begin{tabular}[c]{@{}c@{}}Image \\ Variation~\cite{sdvariation}\end{tabular} & SJC-I~\cite{wang2022score}    & \multicolumn{1}{c}{Ours} \\
    \midrule
    PSNR $\uparrow$ & \underline{8.933}    & 5.914           & 6.573  & \textbf{18.378}      \\ 
    SSIM ~$\uparrow$ & \underline{0.645}    & 0.540           & 0.552  & \textbf{0.877}       \\ 
    LPIPS $\downarrow$ & \underline{0.412}    & 0.545           & 0.484  & \textbf{0.088}       \\ 
    FID ~~~~$\downarrow$ & \underline{12.919}   & 22.533          & 19.783 & \textbf{0.027}       \\ \bottomrule
    \end{tabular}
    \caption{
    \textbf{Results for novel view synthesis on Google Scanned Objects.}
    % Each object is rendered under 10 random views.
    % todo discussion \PT{todo emphasize we are much better than baselines on GSO}
    All metrics demonstrate that our method is able to outperform the baselines by a significant margin.
    \vspace{-0.2cm}
    }
    \label{tab:GSO_NVS}
\end{table}

For novel view synthesis, we numerically evaluate our method and baselines extensively with four metrics covering different aspects of image similarity: PSNR, SSIM~\cite{wang2004image}, LPIPS~\cite{zhang2018unreasonable}, and FID~\cite{heusel2017gans}.
For 3D reconstruction, we measure Chamfer Distance and volumetric IoU. % to evaluate the reconstructed geometry in multiple aspects.

\subsection{Novel View Synthesis Results}
\label{sec:nvs}

We show the numerical results in Tables~\ref{tab:GSO_NVS} and~\ref{tab:RTMV_NVS}.
% \BV{todo maybe extra discussion beyond caption?}
% \paragraph{Qualitative results.}
Figure~\ref{fig:qualitative_gso} shows that our method, as compared to all baselines on GSO, is able to generate highly photorealistic images that are closely consistent with the ground truth. Such a trend can also be found on RTMV in Figure~\ref{fig:qualitative_rtmv}, even though the scenes are out-of-distribution compared to the Objaverse dataset. Among our baselines, we observed that Point-E tends to achieve much better results than other baselines, maintaining impressive zero-shot generalizability. However, the small size of the generated point clouds greatly limits the applicability of Point-E for novel view synthesis.

\begin{table}[t!]
    \footnotesize
    \centering
    \begin{tabular}{cccc|c}
    \toprule
    \multicolumn{1}{l}{} & DietNeRF~\cite{jain2021putting} & \begin{tabular}[c]{@{}c@{}}Image \\ Variation~\cite{sdvariation}\end{tabular} & SJC-I~\cite{wang2022score}    & \multicolumn{1}{c}{Ours} \\ \midrule
    PSNR $\uparrow$ & 7.130    & 6.561      & \underline{7.953}  & \textbf{10.405}          \\
    SSIM ~$\uparrow$ & 0.406    & 0.442      & \underline{0.456}  & \textbf{0.606}           \\
    LPIPS $\downarrow$ & \underline{0.507}    & 0.564      & 0.545  & \textbf{0.323}           \\
    FID ~~~~$\downarrow$ & \underline{5.143}    & 10.218     & 10.202 & \textbf{0.319}           \\ \bottomrule
    \end{tabular}
    \caption{
    \textbf{Results for novel view synthesis on RTMV.}
    Scenes in RTMV are out-of-distribution from Objaverse training data, yet our model still outperforms the baselines by a significant margin.
    \vspace{-0.2cm}
    }
    \label{tab:RTMV_NVS}
\end{table}

\begin{figure*}
    \centering
    \includegraphics[width=\linewidth]{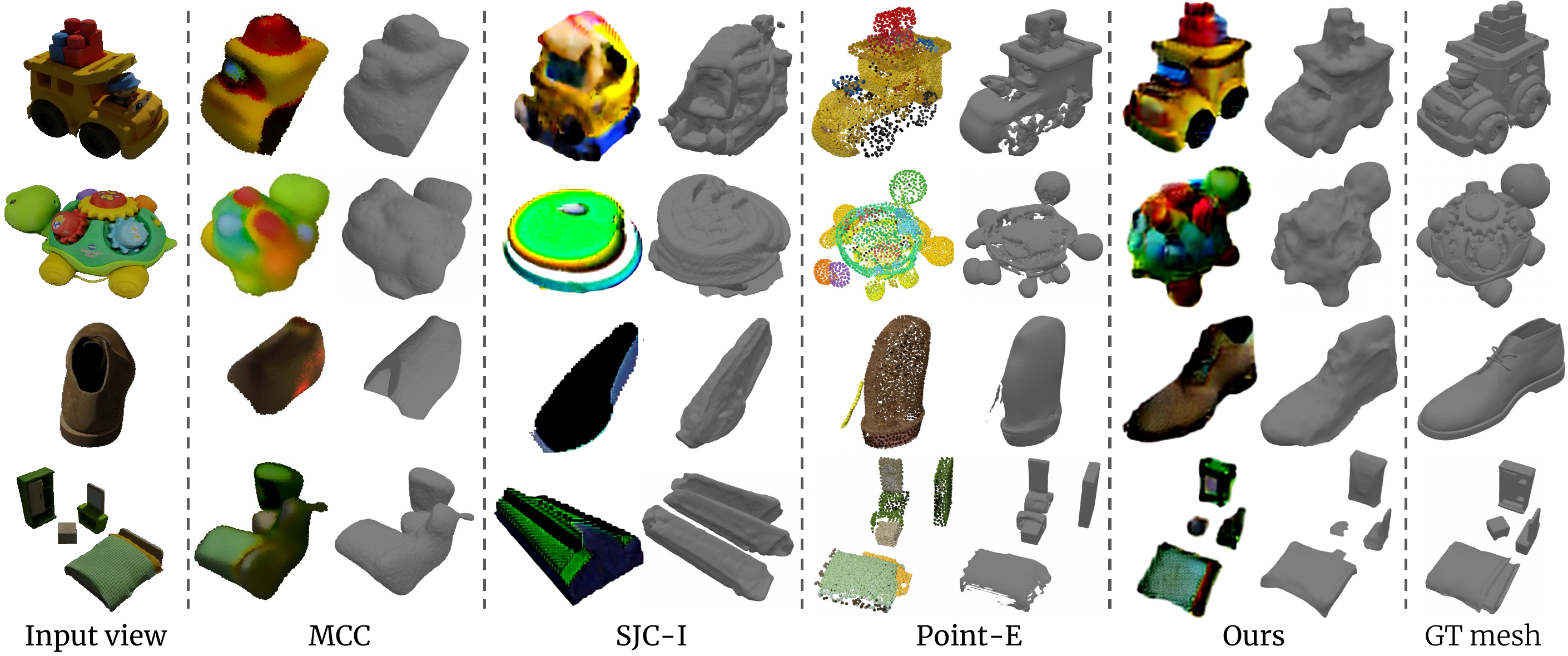}
    \caption{\textbf{Qualitative examples of 3D reconstruction.} The input view images are shown on the left. For each method, we show a rendered view from a different angle and the reconstructed 3D mesh. The ground truth meshes are shown on the right.
    \vspace{-0.5cm}
    }
    \label{fig:geometry}
\end{figure*}

In Figure~\ref{fig:qualitative_wild}, we further demonstrate the generalization performance of our model to objects with challenging geometry and texture as well as its ability to synthesize high-fidelity viewpoints while maintaining the object type, identity and low-level details.
% \BV{is this true? angle is not an input to the VoRF right? don't you just mean lighting intensity variations (which is still lambertian!) instead of specular reflection?}
% \rl{For example, the purple monster in the 3rd row has a near-Lambertian surface while the toy taxi in the 4th row shows some specular reflection. Our model is able to successfully maintain these details in the novel views.}

% \paragraph{Quantitative results.}
% \BV{todo refer to table 1 and table 2}
% All of them highlight that our method is able to outperform the baselines by a significant margin.

%\vspace{+0.5cm}
\textbf{Diversity across samples.}
Novel view synthesis from a single image is a severely under-constrained task, which makes diffusion models a particularly apt choice of architecture compared to NeRF in terms of capturing the underlying uncertainty.
% \BV{I disagree with analogy because this is an essential feature of diffusion models in general, not just text2image}
% Analogy can be drawn here with the text-to-image generation task, in which diffusion models excel at generating high-fidelity.
Because input images are 2D, they always depict only a partial view of the object, leaving many parts unobserved. Figure~\ref{fig:diversity} exemplifies the diversity of plausible, high-quality images sampled from novel viewpoints.
% and all outputs are clearly plausible the is All of them are clearly plausible

\subsection{3D Reconstruction Results}
\label{sec:3d}

We show numerical results in Tables~\ref{tab:GSO_3D} and~\ref{tab:RTMV_3D}.
% \paragraph{Qualitative results.}
Figure~\ref{fig:geometry}  qualitatively shows our method reconstructs high-fidelity 3D meshes that are consistent with the ground truth. 
MCC tends to give a good estimation of surfaces that are visible from the input view, but often fails to correctly infer the geometry at the back of the object.

\begin{table}[t!]
    \footnotesize
    \centering
    \begin{tabular}{lccc|c}
    \toprule
        & MCC~\cite{wu2023multiview}    & SJC-I~\cite{wang2022score}    & Point-E~\cite{nichol2022point} & Ours            \\ \midrule
    CD ~$\downarrow$  & 0.1230 & 0.2245 & \underline{0.0804}  & \textbf{0.0717} \\ 
    IoU $\uparrow$  & 0.2343 & 0.1332 & \underline{0.2944}  & \textbf{0.5052} \\ \bottomrule
    \end{tabular}
    \caption{
    \textbf{Results for single view 3D reconstruction on GSO.} 
    Note that our volumetric IoU is better than the compared methods by a large margin.
    % todo discussion
    % on 20 randomly selected objects from the GSO dataset.
    \vspace{-0.2cm}
    }
    \label{tab:GSO_3D}
\end{table}

SJC-I is also frequently unable to reconstruct a meaningful geometry.
On the other hand, Point-E has an impressive zero-shot generalization ability, and is able to predict a reasonable estimate of object geometry. However, it generates non-uniform sparse point clouds of only 4,096 points, which sometimes leads to holes in the reconstructed surfaces (according to their provided mesh conversion method). % (we use their default meshing method). 
Therefore, it obtains a good CD score but falls short of the volumetric IoU.
% In addition, the 3D voxel NeRF we reconstructed is much higher resolution and scalable than Point-E, which generates 4,000 colored point clouds, regardless of the complexity of the scenes. 
Our method leverages the learned multi-view priors from our view-conditioned diffusion model and combines them with the advantages of a NeRF-style representation.
Both factors provide improvements in terms of CD and volumetric IoU over prior works, as indicated by Tables~\ref{tab:GSO_3D} and \ref{tab:RTMV_3D}.

\begin{table}[t!]
    \footnotesize
    \centering
    \begin{tabular}{lccc|c}
    \toprule
        & MCC~\cite{wu2023multiview}    & SJC-I~\cite{wang2022score}    & Point-E~\cite{nichol2022point} & Ours            \\ \midrule
    CD ~$\downarrow$   & 0.1578 & \underline{0.1554} & 0.1565  & \textbf{0.1352} \\ 
    IoU $\uparrow$  & \underline{0.1550} & 0.1380 & 0.0784  & \textbf{0.2196} \\ \bottomrule
    \end{tabular}
    \caption{
    \textbf{Results for single view 3D reconstruction on RTMV.}
    Because RTMV consists of cluttered scenes with many objects in them, none of the studied approaches seem to perform very well. Our method is still the best one however, despite not being explicitly trained for the 3D reconstruction task.
    % 10 scenes
    \vspace{-0.2cm}
    }
    \label{tab:RTMV_3D}
\end{table}

\begin{figure*}
    \centering
    % \vspace{-0.5em}
    \includegraphics[width=\linewidth]{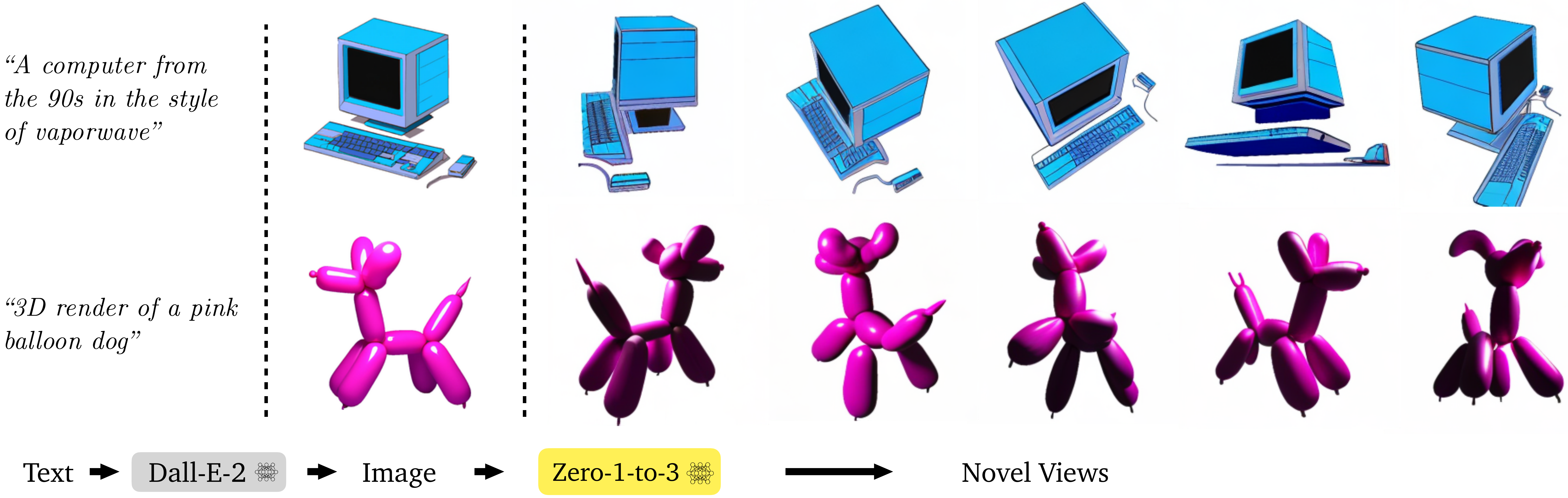}
    \caption{\textbf{Novel View Synthesis from Dall-E-2 Generated Images.} The composition of multiple objects (1st row) and the lighting details (2nd row) are preserved in our synthesized novel views.}
    % \vspace{-1em}
    \label{fig:txt2img}
\end{figure*}

\subsection{Text to Image to 3D} \label{supp:aigc}
In addition to in-the-wild images, we also tested our method on images generated by txt2img models such as Dall-E-2~\cite{ramesh2021zero}. As shown in Figure~\ref{fig:txt2img}, our model is able to generate novel views of these images while preserving the identity of the objects. We believe this could be very useful in many text-to-3D generation applications.

\section{Discussion}
In this work, we have proposed a novel approach, \textit{Zero-1-to-3}, for zero-shot, single-image novel-view synthesis and 3D reconstruction. Our method capitalizes on the Stable Diffusion model, which is pre-trained on internet-scaled data and captures rich semantic and geometric priors. To extract this information, we have fine-tuned the model on synthetic data to learn control over the camera viewpoint. The resulting method demonstrated state-of-the-art results on several benchmarks due to its ability to leverage strong object shape priors learned by Stable Diffusion.

\subsection{Future Work}
%\vspace{-7px}
\textbf{From objects to scenes.} Our approach is trained on a dataset of single objects on a plain background. While we have demonstrated a strong degree of generalizations to scenes with several objects on RTMV dataset, the quality still degrades compared to the in-distribution samples from GSO. Generalization to scenes with complex backgrounds thus remains an important challenge for our method. 

%\vspace{-7px}
\textbf{From scenes to videos.} Being able to reason about geometry of dynamic scenes from a single view would open novel research directions, such as understanding occlusions~\cite{revealing, liu2022shadows} and dynamic object manipulation. A few approaches for diffusion-based video generation have been proposed recently~\cite{ho2022video,esser2023structure}, and extending them to 3D would be key to opening up these opportunities.

%\vspace{-7px}
\textbf{Combining graphics pipelines with Stable Diffusion.} In this paper, we demonstrate a framework to extract 3D knowledge of objects from Stable Diffusion. A powerful natural image generative model like Stable Diffusion contains other implicit knowledge about lighting, shading, texture, etc. Future work can explore similar mechanisms to perform traditional graphics tasks, such as scene relighting.

{\textbf{Acknowledgements:} We would like to thank Changxi Zheng and Samir Gadre for their helpful feedback. We would also like to thank the authors of SJC~\cite{wang2022score}, NeRDi~\cite{deng2022nerdi}, SparseFusion~\cite{zhou2022sparsefusion}, and Objaverse~\cite{deitke2022objaverse} for their helpful discussions. This research is based on work partially supported by the Toyota Research Institute, the DARPA MCS program under Federal Agreement No.\ N660011924032, and the NSF NRI Award \#1925157.}

% \vspace{-7px}
% \paragraph{Inference speed.} As our approach relies on Stable Diffusion for novel view synthesis, it inherits not only its strong object shape priors, but also its relatively slow inference speed. This limitation will become especially important when applying the method to videos with hundreds of frames. Developing novel techniques to speed-up the diffusion process will thus be critical for scaling diffusion-based reconstruction techniques in the future.

{\small
\bibliographystyle{ieee_fullname}
\bibliography{_main}
}

% SUPP
\newpage

\twocolumn[
\centering
\Large
\textbf{Appendix} \\
\vspace{+1em}
] %< twocolumn

\appendix

\section{Coordinate System \& Camera Model} \label{supp:camera}

\begin{figure}[h]
    \centering
    \vspace{-1em}
    \includegraphics[width=0.5\linewidth]{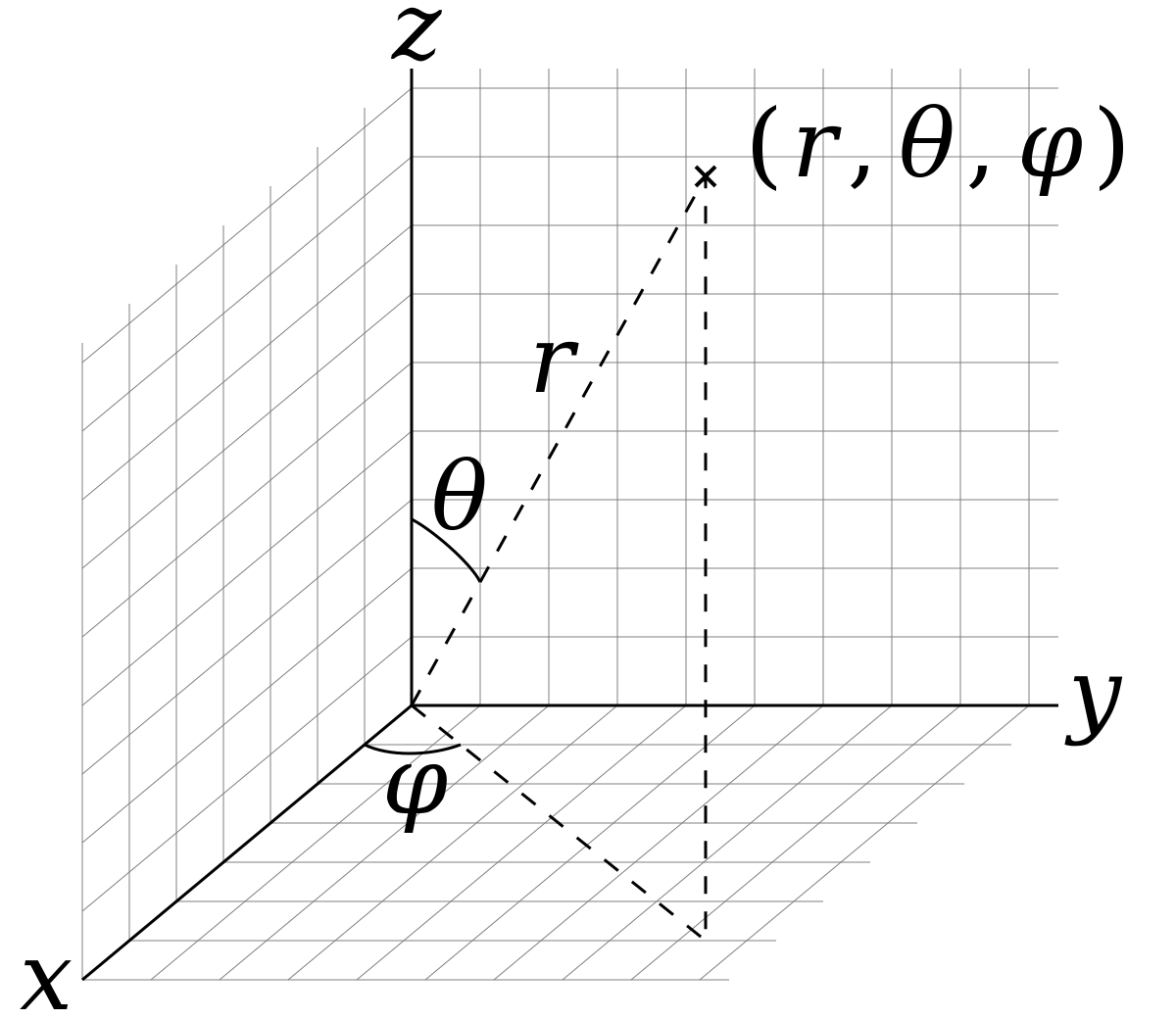}
    \caption{Spherical Coordinate System~\cite{wiki:Spherical_coordinate_system}.}
    \vspace{-1em}
    \label{fig:spherical}
\end{figure}

% We use spherical coordinate systems to represent camera locations and relative transformation and a pinhole camera model. As shown in figure~\ref{fig:spherical}, assuming the center of the object is the origin of the coordinate system, we can use $\theta$, $\phi$, $r$ to represent the polar angle, azimuth angle, and distance from the center. During dataset creation, we normalize all assets to be contained inside a unit cube, then we uniformly sample camera viewpoints where $\theta \sim [0, \pi]$, $\phi \sim [0, 2\pi]$, $r \sim [1.5, 2.2]$.
% NOTE: I also modified this passage but didn't have track changes
We use a spherical coordinate system to represent camera locations and their relative transformations.
As shown in Figure~\ref{fig:spherical}, assuming the center of the object is the origin of the coordinate system, we can use $\theta$, $\phi$, and $r$ to represent the polar angle, azimuth angle, and radius (distance away from the center) respectively. For the creation of the dataset, we normalize all assets to be contained inside the XYZ unit cube $[-0.5, 0.5]^3$. Then, we sample camera viewpoints such that $\theta \in [0, \pi]$, $\phi \in [0, 2\pi]$ uniformly cover the unit sphere, and $r$ is sampled uniformly in the interval $[1.5, 2.2]$.
During training, when two images from different viewpoints are sampled, let their camera locations be $(\theta_1, \phi_1, r_1)$ and $(\theta_2, \phi_2, r_2)$. We denote their \emph{relative} camera transformation as $(\theta_2 - \theta_1, \phi_2 - \phi_1, r_2 - r_1)$. Since the camera is always pointed at the center of the coordinate system, the extrinsics matrices are uniquely defined by the location of the camera in a spherical coordinate system. We assume the horizontal field of view of the camera to be $49.1^\circ$, and follow a pinhole camera model.

Due to the incontinuity of the azimuth angle, we encode it with $\phi \mapsto [\sin(\phi), \cos(\phi)]$.
% Subsequently, during training time, four values representing the relative camera viewpoint change, $[\theta, \sin(\phi), \cos(\phi), r]$ are fed to the model. During inference time, an input view and a user-defined viewpoint change vector $[\theta, \phi, r]$ are used to generate a novel view.
Subsequently, at both training and inference time, four values representing the relative camera viewpoint change, $[\theta, \sin(\phi), \cos(\phi), r]$ are fed to the model, along with an input image, in order to generate the novel view.

\section{Dataset Creation} \label{supp:dataset}
We use Blender~\cite{blender} to render training images of the finetuning dataset. The specific rendering code is inherited from a publicly released repository\footnote{\url{https://github.com/allenai/objaverse-rendering}} by authors of Objaverse~\cite{deitke2022objaverse}. For each object in Objaverse, we randomly sample 12 views and use the Cycles engine in Blender with 128 samples per ray along with a denoising step to render each image. We render all images in 512$\times$512 resolution and pad transparent backgrounds with white color. We also apply randomized area lighting. In total, we rendered a dataset of around 10M images for finetuning.

\section{Finetuning Stable Diffusion} \label{supp:diffusion}

We use the rendered dataset to finetune a pretrained Stable Diffusion model for performing novel view synthesis. Since the original Stable Diffusion network is not conditioned on multimodal text embeddings, the original Stable Diffusion architecture needs to be tweaked and finetuned to be able to take conditional information from an image. This is done in~\cite{sdvariation}, and we use their released checkpoints. To further adapt the model to accept conditional information from an image along with a relative camera pose, we concatenate the image CLIP embedding (dimension 768) and the pose vector (dimension 4) and initialize another fully-connected layer (772 $\mapsto$ 768) to ensure compatibility with the diffusion model architecture. The learning rate of this layer is scaled up to be 10$\times$ larger than the other layers. The rest of the network architecture is kept the same as the original Stable Diffusion.

\subsection{Training Details}

We use AdamW~\cite{loshchilov2017decoupled} with a learning rate of $10^{-4}$ for training. First, we attempted a batch size of 192 while maintaining the original resolution (image dimension $512 \times 512$, latent dimension $64 \times 64$) for training. However, we discovered that this led to a slower convergence rate and higher variance across batches. Because the original Stable Diffusion training procedure used a batch size of 3072, we subsequently reduce the image size to $256 \times 256$ (and thus the corresponding latent dimension to $32 \times 32$), in order to be able to increase the batch size to 1536. This increase in batch size has led to better training stability and a significantly improved convergence rate. We finetuned our model on an 8$\times$A100-80GB machine for 7 days.

\subsection{Inference Details}
To generate a novel view, Zero-1-to-3 takes only 2 seconds on an RTX A6000 GPU. Note that in prior works, typically a NeRF is trained in order to render novel views, which takes significantly longer. In comparison, our approach inverts the order of 3D reconstruction and novel view synthesis, causing the novel view synthesis process to be fast and contain diversity under uncertainty.  Since this paper addresses the problem of a single image to a 3D object, when an in-the-wild image is used during inference, we apply an off-the-shelf background removal tool~\cite{ophoperhpo} to every image before using it as input to Zero-1-to-3.

% \subsection{Scalability}
% There are several ways to further scale up our method. First, the \textit{Objaverse} dataset continues to grow in size. By training a model on a bigger dataset of 3D assets using our method, we can expect the model to further improve in generalization performance to in-the-wild images. Second, as limited by the resources we have, we can only afford to render 12 views for each object. Therefore, another way to scale up the complexity and diversity of the dataset is to sample significantly more views, e.g. 128 views per object, or even on-the-fly rendering (new views are sampled for each training epoch). Third, we reduced the image and latent resolution during training due to limited resources. We believe a higher image and latent resolution while keeping the same batch size can lead to better results both qualitatively and quantitatively.

\section{3D Reconstruction} \label{supp:3d}

Different from the original Score Jacobian Chaining (SJC) implementation, we removed the ``emptiness loss'' and ``center loss''. To regularize the VoxelRF representation, we differentiably render a depth map, and apply a smoothness loss to the depth map. This is based on the prior knowledge that the geometry of an object typically contains less high-frequency information than its texture. It is particularly helpful in removing holes in the object representation. We also apply a near-view consistency loss to regularize the difference between an image rendered from one view and another image rendered from a nearby randomly sampled view. We found this to be very helpful in improving the cross-view consistency of an object's texture.
All implementation details can be found in the code that is submitted as part of the appendix. Running a full 3D reconstruction on an image takes around 30 minutes on an RTX A6000 GPU.

\paragraph{Mesh extraction.}
We extract the 3D mesh from the VoxelRF representation as follows.
We first query the density grids at resolution $200^3$.
Then we smooth the density grids using a mean filter of size $(7, 7, 7)$, followed by an erosion operator of size $(5, 5, 5)$.
Finally, we run marching cubes on the resulting density grids.
Let $\bar d$ denote the average value of the density grids.
For the GSO dataset, we use a density threshold of $8\bar d$.
For the RTMV dataset, we use a density threshold of $4\bar d$.

\paragraph{Evaluation.} 
The ground truth 3D shape and the predicted 3D shape are first normalized within the unit cube. 
To compute the chamfer distance (CD), we randomly sample 2000 points. 
For Point-E and MCC, we sample from their predicted point clouds directly.
For our method and SJC-I, we sample points from the reconstructed 3D mesh.
We compute the volumetric IoU at resolution $64^3$.
For our method, Point-E and SJC-I, we vocalize the reconstructed 3D surface meshes using marching cubes.
For MCC, we directly voxelize the predicted dense point clouds by occupancy.

\section{Baselines} \label{supp:baseline}

To be consistent with the scope of our method, we compare only to methods that (1) operate in a zero-shot setting, (2) use single-view RGB images as input, and (3) have official reference implementations available online that can be adapted in a reasonable timeframe.
In the following sections, we describe the implementation details of our baselines.

\subsection{DietNeRF}

We use the official implementation located on 
GitHub\footnote{\url{https://github.com/ajayjain/DietNeRF}}, which, at the time of writing, has code for low-view NeRF optimization from scratch with a joint MSE and consistency loss, though provides no functionality related to finetuning PixelNeRF. For fairness, we use the same hyperparameters as the experiments performed with the NeRF synthetic dataset in~\cite{jain2021putting}. For the evaluation of novel view synthesis, we render the resulting NeRF from the designated camera poses in the test set.
% treat the predictions from the test set viewpoints in exactly the same way as the evaluations of Zero-1-to-3.

\subsection{Point-E}

We use the official implementation and pretrained models located on GitHub\footnote{\url{https://github.com/openai/point-e}}.
We keep all the hyperparameters and follow their demo example to do 3D reconstruction from single input image. The prediction is already normalized, so we do not need to perform any rescaling to match the ground truth.
For surface mesh extraction, we use their default method with a grid size of 128.

\subsection{MCC}

We use the official implementation located on 
GitHub\footnote{\url{https://github.com/facebookresearch/MCC}}. Since this approach requires a colorized point cloud as input rather than an RGB image, we first apply an online off-the-self foreground segmentation method~\cite{ophoperhpo} as well as a state-of-the-art depth estimation method~\cite{ranftl2020towards, ranftl2021vision} for preprocessing. For fairness, we keep all hyperparameters the same as the zero-shot, in-the-wild experiments described in~\cite{wu2023multiview}. For the evaluation of 3D reconstruction, we normalize the prediction, rotate it according to camera extrinsics, and compare it with the 3D ground truth.
% \BV{todo elaborate on this -- not entirely sure how our own model evaluation plays into this, and how mesh vs point cloud vs voxels works and should be handled here}
% \RW{I describe this in sec F.}

% {\small
% \bibliographystyle{ieee_fullname}
% \bibliography{supp_bib}
% }

% {\small
% \bibliographystyle{ieee_fullname}
% \bibliography{_bib}
% }

\end{document}